\documentclass[11pt]{article}
\usepackage[preprint]{acl}
\usepackage{times}
\usepackage{latexsym}
\usepackage[T1]{fontenc}
\usepackage[utf8]{inputenc}
\usepackage{microtype}
\usepackage{booktabs}
\usepackage{enumitem}
\usepackage{graphicx}
\usepackage{amsmath}
\usepackage{amssymb}
\usepackage{pifont}
\usepackage{xcolor}
\usepackage{float}
\title{What Survives When You Compress a Recursive Reasoner for the Edge?}

\author{
 \textbf{Pearse Jim\textsuperscript{1}\thanks{\, Equal Contribution.}},
 \textbf{Steven Kolawole\textsuperscript{2,1}\footnotemark[1]},
 \textbf{Opegbemi M. Busoye\textsuperscript{1}},
 \textbf{Glory Bagai\textsuperscript{1}},
 \textbf{Virginia Smith\textsuperscript{2}}
\\
\\
 \textsuperscript{1}ML Collective,
 \textsuperscript{2}Carnegie Mellon University
}
\begin{document}
\maketitle

\begin{abstract}
Recursive reasoning models solve hard structured tasks with a few million
parameters by iterating a latent state, but deploying them on edge hardware
means compressing them – and quantization noise compounds across recursive
cycles rather than accumulating over output tokens, so single-pass intuitions
fail. Here, we ask what survives. Across a full precision sweep and three
tasks, aggressive compression preserves \emph{local} prediction but destroys
\emph{global} reasoning: pruning, distillation, and linear attention drive
puzzle-exact accuracy to zero while cell accuracy holds. Na\"ive INT4 is
different: its damage is architectural, striking MLP-mixing recursion but not
attention on the same task, and per-channel calibration reverses it without
retraining, where quantization-aware training does not. We also introduce
carry-trajectory fidelity, the cosine similarity to the full-precision
reasoning path, as a label-free signal that predicts both the damage and the
recovery before any task evaluation, with a threshold that transfers across
tasks. The result is a deployment recipe: flash-streamed embeddings remove a
99.4\,MB bottleneck, one recursive cycle matches full-depth accuracy at
$6\times$ fewer FLOPs, and calibrated INT4 brings the backbone to 3.3\,MB,
inside a 4\,MB budget.

\end{abstract}

% ============================================================
% Introduction Section
% TRM Quantization Paper — EMNLP 2026 Industry Track
% ============================================================

\begin{figure}[t]
%\vspace{-3mm}
\centering
\includegraphics[width=0.90\columnwidth]{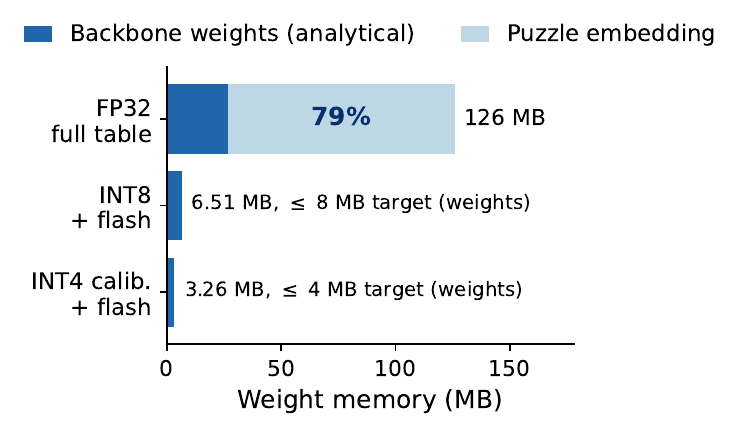}
%\vspace{-3mm}
\caption{Backbone-weight memory by configuration (analytical, weights-only).
The FP32 embedding table (99.4\,MB) accounts for 79\% of the $\sim$126\,MB
baseline. Flash loading reduces the active embedding to 2\,KB, isolating
backbone compression. INT8 + flash reaches 6.51\,MB and calibrated INT4 + flash 3.26\,MB, the
latter within the 4\,MB microcontroller budget. Both are weights-only; with
activations and carry state the ARC/Maze streaming peak is 9.51\,MB
(Table~\ref{tab:memfigures}).}
\label{fig:memory}
%\vspace{-5mm}
\end{figure}

%\vspace{-2mm}
\section{Introduction}
%\vspace{-2mm}

The Tiny Recursive Model (TRM) solves abstract reasoning puzzles with only
6.83M parameters~\citep{fan2025trm}---44.6\% on ARC-AGI-1 with 1000-augmentation
test-time voting (36.00\% single-pass in our harness)---competitive with larger autoregressive
systems on a benchmark designed to resist pattern matching~\citep{chollet2019arc}.
It does so by iterating a latent carry state across recursive cycles, a mechanism shared by Universal
Transformers~\citep{dehghani2019universal} and depth-recurrent
language models~\citep{geiping2025recurrentdepth, wang2025hrm, hao2024coconut, kohli2026loopthink}.
Yet TRM weighs $\sim$126\,MB, more than an order of magnitude beyond the
kB-to-MB SRAM of the microcontroller tier we target. Deployment there requires
compression. When such a model is squeezed, what survives and what breaks?

While single-pass transformer quantization is well-studied (noise accumulates over output tokens, and INT8 deployment is standard)~\citep{nagel2021whitepaper, dettmers2022llmint8, han2016deepcompression}, recursive models introduce a different failure mode: quantization noise compounds across $H$ recursive cycles at each carry-state update. Whether this compounding degrades reasoning has not been studied. While quantization harms single-pass reasoning chains~\citep{liu2025quantizationreasoning} and emergent abilities collapse below 4-bit~\citep{liu2024emergent}, depth and precision are independent in those systems. In recursive models, they are not: the optimal operating point depends jointly on bit-width \emph{and} depth.

We present the first systematic compression study of a recursive reasoning model under hard SRAM constraints~\citep{lin2020mcunet, abushahla2025mcusurvey}. We sweep precision (FP32, FP16, INT8, INT4) and depth across three tasks (ARC-AGI-1, Maze-Hard, Sudoku-Extreme) and two architectures (TRM and HRM). As shown in Figure~\ref{fig:memory}, the dominant memory bottleneck is the 99.4\,MB puzzle embedding table, which we resolve via 2\,KB flash loading, isolating backbone compression. Our primary contributions are:
%\vspace{-2mm}

\begin{enumerate}[leftmargin=*]
\itemsep0em 
\item We establish a \textbf{depth--precision crossover} with two separable
  effects: a single recursive cycle matches full-depth accuracy at $6\times$
  fewer FLOPs---a depth effect (FP32 at $H{=}1$ scores 35.75\% vs.\ INT8's
  35.25\%)---while INT8 adds a $4\times$ weight-memory reduction at no accuracy
  cost, an interaction absent from single-pass quantization (\S\ref{sec:int8}).
\item We identify a \textbf{compositional collapse}: under pruning,
  distillation, and linear-attention substitution, local token accuracy
  survives while global puzzle accuracy falls to zero; na\"ive INT4's harm is
  architecture-dependent, and an attention-vs-MLP-mixing ablation shows the
  fragility is architectural, not task-intrinsic (\S\ref{sec:archsensitivity}).
\item We introduce \textbf{carry-trajectory fidelity}, a label-free detector
  that grades this damage and predicts its recovery from the model's own hidden
  states, with no task labels (\S\ref{sec:diagnostic}).
\item We derive and benchmark a \textbf{deployment recipe}: per-channel
  calibrated INT4 weights reach 3.26\,MB, within a 4\,MB microcontroller
  budget, with latency, memory, and file size measured directly on three
  Snapdragon mobile platforms
  (\S\ref{sec:recovery},~\ref{sec:deployment}).
\end{enumerate}

% ============================================================
% Related Work Section — argument-driven
% TRM Quantization Paper — EMNLP 2026 Industry Track
% ============================================================
%\vspace{-3mm}
\section{Related Work}

%\vspace{-2mm}
\paragraph{Overthinking in recursive models is named but not detectable.}
Recurrent and weight-tied architectures are a parameter-efficient path to
algorithmic reasoning~\citep{dehghani2019universal}, and recent depth-recurrent
language models show iterative latent refinement scales
favorably~\citep{geiping2025recurrentdepth, wang2025hrm, hao2024coconut,
kohli2026loopthink}. \citet{bansal2022overthinking} name their central
failure---\textit{overthinking}, where too many iterations corrupt the
instance---but detect it only by full task evaluation;
\citet{du2025latentthinking} and \citet{lu2025latentcot} find correctness
signals in latent trajectories, yet none give a signal usable \emph{during
inference} without labels. Our carry-trajectory fidelity, grounded in
\citet{jiang2025tracing}'s link between hidden-state cosine similarity and
representational saturation, predicts collapse under quantization with no task
evaluation.

%\vspace{-1mm}
\paragraph{Quantization-as-regularization has an uncharacterized depth
dimension.}
Quantization noise can act as a regulariser – pushing toward flatter minima
with a model-specific optimal level~\citep{askari2024qgen, ravula2022qreg},
robust under distribution shift and limited data~\citep{javed2025qtdog,
ferianc2023fightingoverfitting}, sometimes letting INT8 beat
FP32~\citep{liao2024quantcontinual} -- which explains INT8 parity in single-pass
models. None predicts recursive behavior, where the same noise is applied
across $H$ carry-state updates and beneficial regularization at shallow depth
becomes destructive compounding at depth. We characterize this depth--precision
interaction directly.

%\vspace{-1mm}
\paragraph{Compressed reasoning fails early; low-bit recovery needs the right
recipe.}
On reasoning chains, PTQ induces early failures that
cascade~\citep{liu2025quantizationreasoning}, with a precision floor near
4-bit~\citep{liu2024emergent} and W8A8 as the lossless threshold for small
models~\citep{liu2025quanthurts}; consistent with our INT8 finding, but for
single-pass decoding. Standard QAT recipes for reasoning models favor
distillation/KD objectives and PTQ initialization~\citep{lv2025lowbitqat}; our
QAT comparison follows that recipe (calibrated initialization, CE and KD
objectives, three learning rates) and finds that further token-level training
degrades an already-calibrated model (\S\ref{sec:qat}), so the quantizer is the
reliable recovery path in our setting.

%\vspace{-1mm}
\paragraph{Edge NLP deployment motivates but does not resolve these questions.}
Foundational compression~\citep{han2016deepcompression}, the finding that
pruning degrades small LMs more than quantization~\citep{zhou2025pruningquant}
(consistent with our pruning results), INT8 as the deployment-native
format~\citep{dettmers2022llmint8, kim2021ibert}, sub-megabyte MCU inference via
hardware co-design~\citep{lin2020mcunet}, MCU-quantization surveys at our
4--8\,MB constraints~\citep{abushahla2025mcusurvey}, and on-device NLP
pipelines~\citep{bohdal2025ondevice}: none studies recursive reasoning models or
the depth--precision interplay. No prior work compresses a recursive reasoning
model; the closest~\citep{liu2025quantizationreasoning} addresses single-pass
decoding and lacks the recursive carry-state structure we study.

% ============================================================
% Background + Experimental Setup
% TRM Quantization Paper — EMNLP 2026 Industry Track
% ============================================================
%\vspace{-2mm}
\section{Background and Experimental Setup}
\label{sec:methods}

%\vspace{-2mm}
\subsection{Tiny Recursive Model (TRM)}
%\vspace{-1mm}

TRM~\citep{fan2025trm} is a weight-tied recursive transformer that solves
structured reasoning tasks through iterative latent refinement rather than
autoregressive generation. Let $\mathcal{V}$ be a finite vocabulary and
$X \in \mathcal{V}^{T_x}$ be an input sequence. TRM maintains a latent
carry state updated across two nested loops.

Given an initial answer embedding $Y_0 \in \mathbb{R}^{T_y \times d}$,
the model maintains a carry state $Z_h \in \mathbb{R}^{T_z \times d}$
across outer cycles $h \in \{1, \dots, H\}$. At each outer cycle, an
inner loop executes $L$ refinement steps. A shared two-layer transformer
block $f_\theta$ updates the state:
%\vspace{-2mm}
\begin{equation}
    Z_h^{(l)} = f_\theta\!\left([X \,;\, Y_{h-1} \,;\, Z_h^{(l-1)}]\right)
%\vspace{-2mm}
\end{equation}

where $[\cdot;\cdot]$ denotes sequence concatenation. After $L$ inner steps,
the outer loop advances via:
%\vspace{-2mm}
\begin{equation}
    Y_h,\; Z_{h+1}^{(0)} = g_\phi\!\left(Z_h^{(L)}\right)
%\vspace{-2mm}
\end{equation}
A one-step gradient approximation during training keeps memory $\mathcal{O}(1)$
with respect to recursive depth. The total number of refinement steps is
$H \times n_\text{sup}$, where $n_\text{sup}$ is the number of outer-loop
supervision steps. At the default ARC-AGI-1 configuration ($H=3$,
$n_\text{sup}=16$), TRM executes 48 refinement steps,
incurring $\sim$3,000\,GFLOPs per puzzle over a 900-token
sequence ($\mathcal{O}(n^2)$ attention).

We study two architectural variants: \textbf{TRM-Attention}, which uses
multi-head self-attention over the full token sequence, and
\textbf{TRM-MLP-Mixing}, which replaces attention with token-mixing MLP
layers. These variants differ in how relational structure is encoded across
refinement steps, which has consequences for compression sensitivity
(\S\ref{sec:archsensitivity}).

%\vspace{-1mm}
\subsection{Tasks and Datasets}
%\vspace{-1mm}

\paragraph{ARC-AGI-1.} Abstract visual reasoning requiring detection of
transformation rules across grid inputs~\citep{chollet2019arc}. We evaluate
on the ARC-AGI-1 public evaluation set (400 tasks). TRM uses
TRM-Attention with a 50,911-entry puzzle embedding table (99.4\,MB in FP32)
that provides puzzle-specific context and, in its published protocol,
test-time augmentation (1,000 augmentations per puzzle) with multi-pass
majority voting; our evaluations run single-pass (no TTA) unless noted.
Parameters: 6.83M.

\paragraph{Maze-Hard.} Path-finding requiring exact end-to-end traversal
over topological maps. Uses TRM-Attention without an embedding table or TTA.
Parameters: 6.82M.

\paragraph{Sudoku-Extreme.} Constraint satisfaction over $9\times9$ grids
with hard row, column, and block exclusions. Uses TRM-MLP-Mixing without an
embedding table. Parameters: 5.03M.

We report two metrics throughout. \textbf{Cell accuracy} is the fraction
of correct grid cells predictions. \textbf{Puzzle exact match}
is the fraction of puzzles solved with zero errors. These metrics can
diverge substantially: a model may predict nearly correct tokens (high
cell accuracy) while satisfying zero global constraints (zero puzzle
exact), as observed under several compression interventions (\S\ref{sec:qat}).

\subsection{Evaluation Protocol}
\label{sec:protocol}

Headline evaluations use TRM's test-time default: $n_\text{sup}{=}16$, an fp32
forward pass, and random test subsets (Table~\ref{tab:corrected}). We call this
the primary protocol. The wider sweeps and ablations, marked at point of use,
run a lighter secondary protocol ($n_\text{sup}{=}10$, bf16 forward, first-$N$
subsets) whose absolute accuracies sit a few points lower. The two protocols
agree on every ordering this paper reports.

\begin{table}[t]
\centering
\small
\setlength{\tabcolsep}{3pt}
\begin{tabular}{@{}lrrrr@{}}
\toprule
\textbf{Task} & \multicolumn{4}{c}{\textbf{Puzzle-exact \% $\uparrow$}} \\
 & \textbf{FP32} & \textbf{INT8} & \textbf{INT4} & \textbf{INT4 cal.} \\
\midrule
Sudoku, TRM-MLP  & 74.6 & 74.7 & 6.4  & 74.5 \\
Sudoku, TRM-Attn & 73.1 & 72.8 & 68.3 & 72.3 \\
Maze, TRM-Attn   & 87.0 & 87.1 & 86.2 & 86.7 \\
\bottomrule
\end{tabular}
\caption{Puzzle-exact accuracy (\%) at the primary protocol
($n_\text{sup}{=}16$, fp32 forward). Sudoku: random $n{=}2{,}000$ subset (both
architectures on the same subset); Maze: full 1{,}000-maze test set. INT4 is
na\"ive per-tensor; INT4 cal.\ is per-channel calibrated.}
\label{tab:corrected}
\end{table}

At this protocol na\"ive INT4 costs MLP-mixing 68.2\,pp against 4.8\,pp for
attention on the same subset (Table~\ref{tab:corrected}).

\subsection{Deployment Targets}
%\vspace{-1mm}
The compression goal is on-device inference with no cloud dependency under
hard SRAM constraints. Our target is the microcontroller tier: kB-to-MB SRAM,
no DRAM, Cortex-M class, where the deployed state of the art is
1--2\,MB CNNs~\citep{lin2020mcunet} and nothing reasoning-capable is deployed
at all. We use two budgets as reference points on the way there, 8\,MB (mobile
SoC SRAM, e.g.\ Cortex-A55) and 4\,MB (MCU, e.g.\ Cortex-M55); the FP32
baseline ($\sim$126\,MB total: 26.7\,MB backbone + 99.4\,MB embeddings)
exceeds both by an order of magnitude.

Measurement cannot reach the target tier directly: no MCU toolchain runs these
graphs today, so the on-device latency, memory, and model sizes of
\S\ref{sec:deployment} come from three Snapdragon platforms via Qualcomm AI
Hub (App.~\ref{app:latency}), which are mobile-class. Everything we report at
the 4\,MB budget is therefore an analytical, weights-only projection
(Table~\ref{tab:memfigures}).

\subsection{Compression Methods}
%\vspace{-1.5mm}

\paragraph{Post-training quantization (PTQ).}
We apply PTQ directly to the pretrained FP32 checkpoint, without retraining.
\textbf{FP16} halves memory via standard half-precision casting.
\textbf{INT8} uses per-channel scale calibration via \texttt{bitsandbytes}
\citep{dettmers2022llmint8} - the standard format for hardware with integer
acceleration units.
\textbf{INT4 na\"ive} applies per-tensor PTQ without calibration.
\textbf{INT4 calibrated} applies per-channel asymmetric PTQ (min/max range per
output channel), again with no fine-tuning; it recovers most of the accuracy
that na\"ive INT4 destroys (\S\ref{sec:recovery}).

Structured pruning and knowledge distillation were also evaluated. Both
collapse puzzle-level accuracy across all tasks while partially preserving
cell accuracy (\S\ref{sec:qat}). We read this as a consequence of the model's
compactness: at 5--7M parameters there is too little redundancy to absorb
pruning, and a single-pass student cannot capture the recursive computation.

%\vspace{-2mm}
\paragraph{Embedding compression.}
For ARC-AGI-1, we separately evaluate three strategies for the 99.4\,MB
puzzle embedding table: INT8 quantization of the full table, SVD rank-16
factorization, and single-puzzle flash loading (streaming only the 2\,KB
row for the current puzzle at inference time).

\subsection{Cycle Reduction}
%\vspace{-2mm}

$H$ (outer cycles) and $n_\text{sup}$ (the number of outer-loop supervision
steps) together determine total recursion depth: $H = 3$, $n_\text{sup} = 16$
means 48 sequential refinement steps on ARC-AGI-1. We sweep
$(H, n_\text{sup}) \in \{(1, 1\text{--}16),\,(3, 1\text{--}16)\}$
for ARC-AGI-1, and $(H, n_\text{sup}) \in \{(1\text{--}4, 1\text{--}10)\}$
for Maze-Hard and Sudoku-Extreme. FLOPs scale linearly with
total steps: at $H=1$, $n_\text{sup}=8$, the per-puzzle cost falls from
$\sim$3,000\,GFLOPs to $\sim$500\,GFLOPs on ARC-AGI-1.

\subsection{Carry-State Diagnostics}
\label{sec:carrydiag}
%\vspace{-2mm}

We use the recursive carry state to build a label-free signal for reasoning
collapse under compression. Let $Z_h \in \mathbb{R}^{T_z \times d}$ be the
carry state after recursive call $h$, with $z_{h,i}$ its $i$-th token vector.

\paragraph{Consecutive similarity (baseline).}
The cosine similarity between successive carry states,
%\vspace{-3mm}
\begin{equation}
    s_h = \frac{1}{T_z} \sum_{i=1}^{T_z}
          \frac{z_{h,i} \cdot z_{h-1,i}}{\|z_{h,i}\|_2\,\|z_{h-1,i}\|_2},
\end{equation}
measures per-step refinement (low $s_h$) versus convergence to a near-fixed
point (high $s_h$), following \citet{jiang2025tracing}'s finding that
hidden-state cosine similarity tracks representational saturation. We find
this signal is nearly invariant to precision (\S\ref{sec:diagnostic}) and thus
a poor failure detector.

\paragraph{Carry-trajectory fidelity (primary).}
The signal we recommend compares the compressed trajectory to the
full-precision one. Let $Z_H^{q}$ and $Z_H^{\text{fp32}}$ be the final carry
states of the quantized and FP32 models on the same input. Fidelity is
%\vspace{-3mm}
\begin{equation}
    \phi = \frac{1}{T_z}\sum_{i=1}^{T_z}
      \frac{z_{H,i}^{q}\cdot z_{H,i}^{\text{fp32}}}
           {\|z_{H,i}^{q}\|_2\,\|z_{H,i}^{\text{fp32}}\|_2}.
\end{equation}
$\phi$ quantifies how far quantization pushes the reasoning trajectory off the
full-precision path. It requires only the FP32 reference (available before
deployment) and no task labels, and---unlike $s_h$---is graded and monotonic
with the eventual accuracy loss (\S\ref{sec:diagnostic}).

\subsection{Quantization-Aware Training and Recovery}
%\vspace{-2mm}

As a probe of whether a token-level objective can repair compression-induced
reasoning loss, we run QAT from the calibrated-INT4 checkpoint: 1{,}000 steps at
batch 32, under cross-entropy and token-distillation objectives at three
learning rates each, with the deep-supervision ACT loop (\S\ref{sec:qat}).
As the alternative recovery path we apply per-channel calibrated INT4 (a
post-training quantizer with no fine-tuning) and report both its accuracy and
its carry-trajectory fidelity (\S\ref{sec:recovery}).

% ============================================================
% Results Section
% TRM Quantization Paper — EMNLP 2026 Industry Track
% ============================================================

\section{Results}
%\vspace{-3mm}

\subsection{The Embedding Table Dominates the Memory Footprint}

The ARC-AGI-1 puzzle embedding table occupies 99.4\,MB of the $\sim$126\,MB FP32
footprint, against 26.7\,MB for the transformer backbone
(Table~\ref{tab:embedding}). Three strategies were
evaluated for embedding compression.

\begin{table}[t]
%\vspace{-2mm}
\centering
\small
\begin{tabular}{@{}lrl@{}}
\toprule
\textbf{Strategy} & \textbf{SRAM} $\downarrow$ & \textbf{Accuracy} $\uparrow$ \\
\midrule
INT8 quant.\ (full table)   & 24.9\,MB & no loss \\
SVD rank-16 factorization   & 3.2\,MB  & lossy ($\cos\!=\!0.94$) \\
Single-puzzle flash loading & 2\,KB    & no loss \\
\bottomrule
\end{tabular}
%\vspace{-2mm}
\caption{Embedding compression strategies on ARC-AGI-1. Only flash loading
brings the embedding within the deployment budgets without loss; the backbone
and the activation-inclusive streaming peak are accounted separately
(\S\ref{sec:deployment}, Table~\ref{tab:sram_footprints}).}
\label{tab:embedding}
%\vspace{-0mm}
\end{table}

Only flash loading fits the deployment targets while preserving accuracy
(Table~\ref{tab:embedding}): the 2\,KB row for the current puzzle is streamed
from flash at inference time, decoupling the embedding from the backbone
compression problem. Figure~\ref{fig:memory} visualizes the resulting
footprints. With the embedding handled, the backbone remains, and with it two
levers: precision and recursion depth.

\subsection{INT8 with Reduced Recursion Matches Full-Depth FP32}
\label{sec:int8}

\begin{figure}[t]
%\vspace{-4mm}
\centering
\includegraphics[width=0.92\columnwidth]{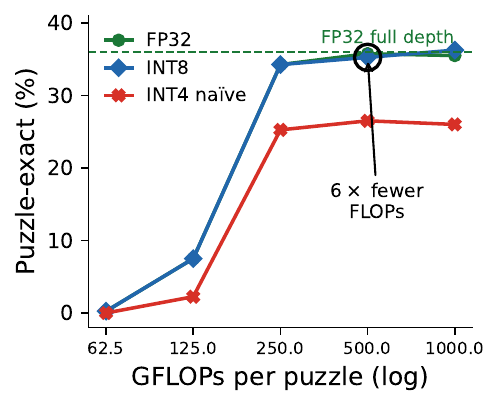}
%\vspace{-3mm}
\caption{ARC-AGI-1 efficiency frontier (single-pass). One recursive cycle
reaches full-depth accuracy at 500\,GFLOPs ($\star$)---a $6\times$ FLOPs
reduction attributable to depth: FP32 at the same depth scores 35.75\%.}
\label{fig:depth}
%\vspace{-7mm}
\end{figure}

INT8 is lossless across both architectural variants and all three tasks
(Table~\ref{tab:full_results}, Appendix~\ref{app:full_results}): per-channel calibration holds quantization
error within what the recursive refinement process tolerates. The key efficiency gain
therefore comes from cycle reduction, not precision: FP32 at
$H{=}1$, $n_\text{sup}{=}8$ already scores 35.75\%
(Table~\ref{tab:arc_sweep_detail}), and INT8 at the same depth achieves
35.25\% puzzle exact on ARC-AGI-1, single-pass (no test-time
augmentation),\footnote{All reported ARC accuracies are single-pass: the
evaluation's fast mode discarded the augmentations, so no augment-and-vote
protocol was applied. TRM's published 44.6\% uses 1000-augmentation TTA and is
not directly comparable. App.~\ref{app:latency} compares the notebook harness
with on-device execution.} within 0.75\,pp of the 36.00\% full-depth baseline at
$6\times$ fewer FLOPs (Figure~\ref{fig:depth}). The two effects are
separable: depth reduction supplies the $6\times$ FLOPs saving, while INT8
supplies a $4\times$ weight-memory reduction at no accuracy cost.

Across the full depth sweep (Appendices~\ref{app:depth_arc}
and~\ref{app:sweeps_detail}), $H{=}1$ brackets the
36.00\% baseline from 34.25\% ($n_\text{sup}{=}4$) to 36.25\%
($n_\text{sup}{=}16$). The reduction holds on Maze-Hard (INT8 at $H{=}1$,
$n_\text{sup}{=}8$ stays within 0.10\,pp at $3.75\times$ fewer FLOPs).
Sudoku-Extreme is the exception: its accuracy keeps climbing with depth
within the swept range (best at $H{=}4$, $n_\text{sup}{\le}10$) and INT8 stays
lossless there, so its full recursion cannot be
cheaply cut (Appendices~\ref{app:depth_maze},~\ref{app:depth_sudoku}).

Dynamic quantization preserves full-precision accuracy on every task, but NPU
accelerators require static, pre-allocated tensor boundaries; the static
path's task-dependent penalties, and the operator exclusions Sudoku requires,
are in Appendix~\ref{app:static_dynamic}. INT8, then, is safe everywhere.
INT4 is not, and where it breaks turns out to be the more informative
finding.

\subsection{Compression Fragility Is Architectural, Not Task-Intrinsic}
\label{sec:archsensitivity}
%\vspace{-2.5mm}
The INT4 damage follows the architecture, not the task: the TRM-Attention
tasks (ARC, Maze) tolerate na\"ive INT4 while TRM-MLP-Mixing (Sudoku)
collapses (Table~\ref{tab:full_results}). This looks like a task-precision
effect; a same-task ablation shows it is not (Figure~\ref{fig:crosstask};
full grid in Appendix~\ref{app:arch_ablation}).

\begin{figure}[t]
%\vspace{-2mm}
\centering
\includegraphics[width=0.92\columnwidth]{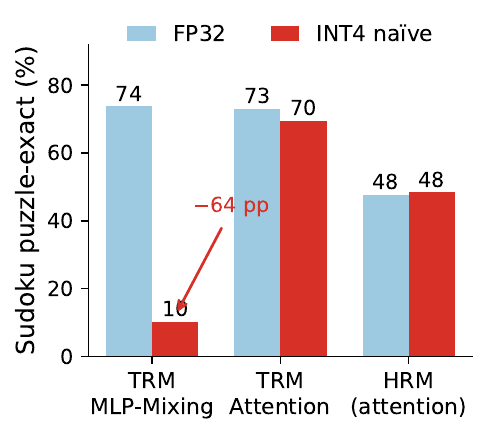}
%\vspace{-5mm}
\caption{Architecture ablation on Sudoku-Extreme: full precision vs.\ na\"ive
INT4 puzzle-exact for three recursive architectures on the \emph{same} task
(secondary protocol; the primary protocol widens the gap to $-68$\,pp,
Table~\ref{tab:corrected}). Only TRM-MLP-Mixing collapses; TRM-Attention (at
the same FP32 accuracy) and HRM (a different model family) survive; the
fragility follows the MLP-mixing token mixer.}
\label{fig:crosstask}
%\vspace{-8mm}
\end{figure}

Under na\"ive INT4, Maze-Hard loses 0.8\,pp and ARC-AGI-1 loses 10.1\,pp puzzle
exact while keeping 89.2\% cell accuracy, but Sudoku-Extreme collapses by
68.2\,pp (primary protocol, Table~\ref{tab:corrected}; the official-checkpoint
ARC figures are in \S\ref{sec:recovery}). Since Sudoku was tested only with
MLP-mixing, we re-run it with attention: TRM-Attention, at nearly the same
full-precision accuracy on the same subset (73.1\% vs.\ 74.6\%), retains 68.3\%
under na\"ive INT4, and HRM~\citep{wang2025hrm} is also robust, while only
MLP-mixing collapses (Table~\ref{tab:arch_ablation}). The accuracy match rules
out a headroom confound, locating the fragility in the token mixer. HRM,
evaluated only on Sudoku and at a lower FP32 baseline, corroborates the pattern
in a second model family rather than serving as a controlled comparison.

\begin{figure}[t]
\centering
\includegraphics[width=\columnwidth]{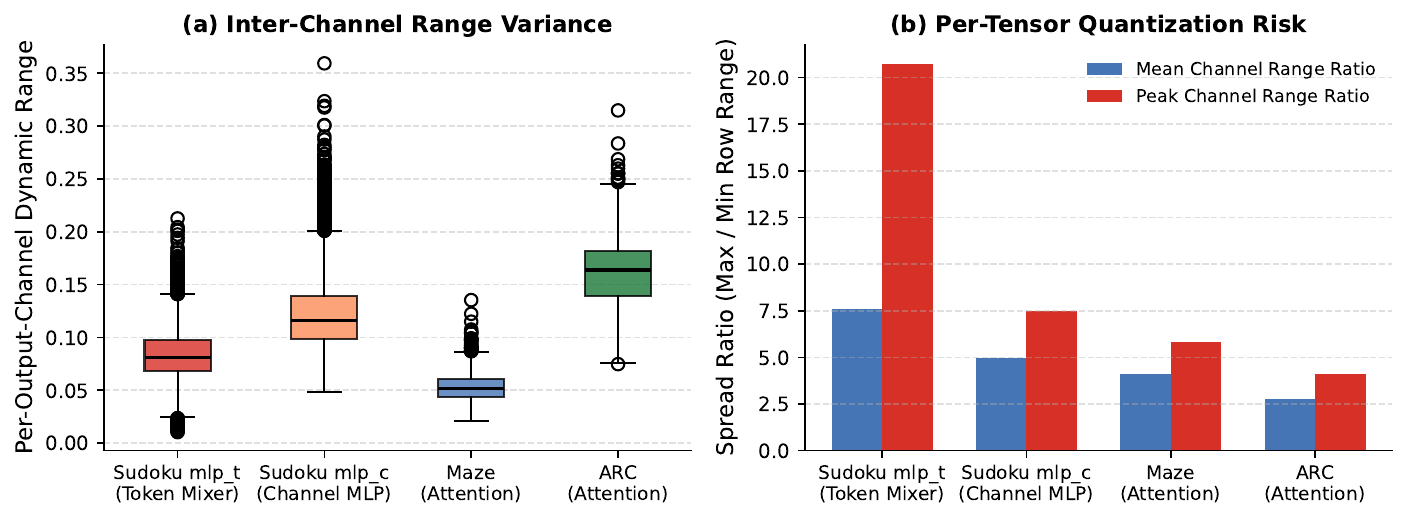}
\caption{Per-output-channel weight range by projection type, across the
deployment models. (a) The distribution for each projection. (b) The ratio of
widest to narrowest channel range --- what a single per-tensor scale must span.
Sudoku's token mixer (\texttt{mlp\_t}) is the outlier; its TRM-Attention
counterpart, used as the same-task control in \S\ref{sec:archsensitivity}, is not
plotted.}
\label{fig:weightrange}
\end{figure}

\paragraph{A weight-level mechanism.}
Per-output-channel dynamic range explains why the token mixer specifically. On
the two Sudoku checkpoints the token-mixing projections reach a 20.7$\times$
inter-channel spread, against 11.1$\times$ for the worst attention projection.
Figure~\ref{fig:weightrange} shows the underlying distributions and extends the
comparison to the deployment models: within the same network the token mixer's
spread is nearly three times the channel MLP's, and both attention models sit
lower still. The token mixer is the outlier on every comparison available to
us, within a network and across them. Under a single per-tensor
INT4 scale, the widest-spread channels dominate the quantization grid and
narrow channels collapse into one or two bins. Per-channel calibration repairs
exactly this, and the recovery of \S\ref{sec:recovery} follows. A failure rooted
in per-channel spread implies a per-channel fix, which the next section tests.

\subsection{Calibrated INT4 Recovers Constraint Reasoning}
\label{sec:recovery}
%\vspace{-2.5mm}

The catastrophic Sudoku-INT4 result is specific to \emph{na\"ive per-tensor}
quantization. Replacing it with per-channel calibrated INT4 (still 4-bit
weights, still post-training, no fine-tuning) recovers essentially all of the
lost accuracy: at the primary protocol, na\"ive INT4 drops Sudoku to 6.4\%
puzzle exact while calibrated INT4 reaches 74.5\%, within 0.1\,pp of the 74.6\%
FP32 baseline on the same subset (Table~\ref{tab:corrected}; secondary-protocol figures in
App.~\ref{app:carry_traj}). The carry-trajectory fidelity diagnostic tracks
this recovery with no task evaluation: fidelity rises from $0.35$ (na\"ive,
broken) to $0.89$ (calibrated, recovered).

\paragraph{Measured on the official ARC checkpoint.}
The same recovery holds on ARC-AGI-1, measured on the official TRM checkpoint
(arcprize verification release; embedding table verified against the rebuilt
official dataset, 876{,}406 identifiers), single-pass over all 400 evaluation
tasks:

\begin{table}[t]
\centering
\small
\setlength{\tabcolsep}{4pt}
\begin{tabular}{@{}lrrrr@{}}
\toprule
\textbf{Variant} & \textbf{pass@1} $\uparrow$ & \textbf{Solved} $\uparrow$ & \textbf{Cell} $\uparrow$ & \textbf{Fidelity} $\uparrow$ \\
\midrule
FP32                & 29.5\% & 115 & 91.5\% & 1.000 \\
INT8 per-channel    & 30.0\% & 117 & 91.6\% & 1.000 \\
INT4 na\"ive        & 19.4\% & 74  & 89.2\% & 0.765 \\
INT4 calibrated     & 29.4\% & 114 & 91.6\% & 0.942 \\
\bottomrule
\end{tabular}
\caption{Official TRM ARC-AGI-1 checkpoint, single-pass, all 400 evaluation
tasks. pass@1 is the ARC per-task score, with partial credit over a task's
multiple test inputs (419 test grids across the 400 tasks); Solved counts tasks
with every test input correct. Calibration recovers ARC to within 0.1\,pp of
FP32; the fidelity column is a fully held-out test of the
\S\ref{sec:diagnostic} threshold, which was never calibrated on ARC or on this
checkpoint.}
\label{tab:arc_official}
\end{table}

Calibration recovers ARC to within 0.1\,pp of FP32 on a checkpoint anyone can
download, and the fidelity column doubles as the held-out threshold test cited
in \S\ref{sec:diagnostic}.

This reframes the deployment floor: for constraint-satisfaction reasoning it is
not INT8 but \emph{calibrated} INT4, provided the quantizer is per-channel.
Na\"ive INT4 remains unusable, and the %fidelity 
diagnostic distinguishes the
two before deployment.

\subsection{Carry-Trajectory Fidelity Predicts Collapse and Recovery}
\label{sec:diagnostic}
%\vspace{-1.5mm}

The damage in \S\ref{sec:archsensitivity} and its repair in
\S\ref{sec:recovery} were both established by running the task. We now ask
whether either can be predicted without doing so: a label-free signal that,
before any task evaluation, tells whether a compression preserves or destroys
reasoning. The obvious
candidate (cosine similarity between consecutive carry states) does not work:
it is task-specific but nearly invariant to precision, changing by
$\le 0.064$ from FP32 to INT4 on every task (App.~\ref{app:carry_traj}), so
it cannot separate harmless from harmful compression.

\begin{figure}[t]
%\vspace{-4mm}
\centering
\includegraphics[width=0.92\columnwidth]{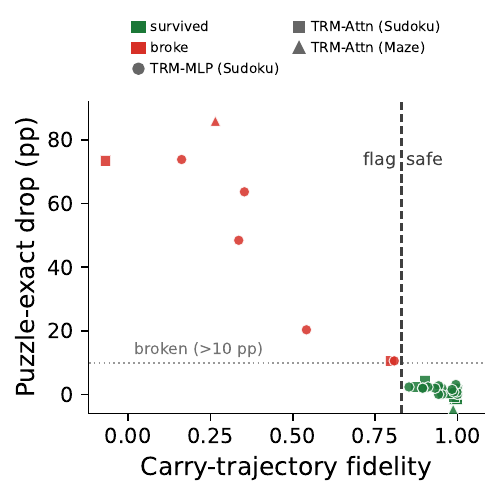}
%\vspace{-4mm}
\caption{Carry-trajectory fidelity is a label-free pass/fail predictor. Each point is one
of the 50 compression configurations in the validation grid (\S\ref{sec:diagnostic}):
bit-widths 3--8, three quantizer granularities and partial-layer sweeps, across three
model--task pairs. Colour marks whether the configuration broke, defined as losing more
than 10\,pp puzzle-exact. The dashed line is the threshold $\tau{=}0.83$ calibrated on
Sudoku alone, which separates the two classes on held-out Maze without error.}
\label{fig:carry}
%\vspace{-5mm}
\end{figure}

The signal that does work is \emph{carry-trajectory fidelity}: the cosine
similarity between the quantized model's final carry state and the
full-precision model's final carry state on the same input
(\S\ref{sec:carrydiag}). It measures how far quantization pushes the reasoning
trajectory off the full-precision path, requires only the FP32 reference
(available before deployment), and uses no task labels. Fidelity is graded and
monotonic with damage (Figure~\ref{fig:carry}; full data Appendix~\ref{app:carry_traj}): it stays $\ge 0.99$ wherever
INT4 is harmless (Maze, and INT8 on every task) and falls in proportion to the
accuracy loss, moderately on ARC and catastrophically on the Sudoku collapse
of \S\ref{sec:archsensitivity} (Table~\ref{tab:carry_fidelity}). It tracks the
calibrated-INT4 \emph{recovery} of \S\ref{sec:recovery} in the same way,
rising as accuracy returns.

\paragraph{The threshold transfers across tasks.}
We validate fidelity as a predictor on a
50-configuration grid spanning bit-widths, quantizer granularities and
partial-layer sweeps across three model--task pairs
(Appendix~\ref{app:carry_traj}); 8 configurations break, defined as a
$>$10\,pp puzzle-exact drop. Spearman $\rho$ between fidelity and accuracy damage is
$-0.75$ overall, fidelity separates broken from surviving configurations
without error on this grid (AUC 1.00), and the dose--response is monotone
(Figure~\ref{fig:carry}). The threshold question now has a held-out answer:
calibrated on Sudoku alone ($\tau=0.83$), it transfers to Maze without error.
The official ARC checkpoint of \S\ref{sec:recovery} is a third, fully
held-out test, on which the threshold was never calibrated: it flags na\"ive
INT4 and clears calibrated INT4 (fidelity column of
Table~\ref{tab:arc_official}). An L2 trajectory-divergence baseline computed
from the same captures is equally predictive on this grid; fidelity's advantage
is being bounded and scale-free across models, not higher discrimination.
Quantization damage is therefore both repairable and predictable in advance.
Neither holds for the remaining compression routes.

\subsection{Cell and Puzzle Accuracy Decouple Under Compression}
\label{sec:qat}
%\vspace{-2mm}

A pattern recurs across our compression interventions: cell accuracy and
puzzle-exact accuracy decouple. A model can predict most individual cells
correctly while satisfying zero global constraints---the signature of a
\emph{compositional-correctness} breakdown that leaves local
token prediction intact. We observe it under three distinct interventions:

On \textbf{knowledge distillation}, a single-cycle student (13--15\% of
  teacher parameters) reaches 87.3\% (Maze) / 54.8\% (Sudoku) cell accuracy
  but 0.0\% puzzle exact (Appendix~\ref{app:pruning_distill},
  Tables~\ref{tab:pruning} and~\ref{tab:distill}).
With \textbf{structured pruning} at 25\% sparsity, cell accuracy partially
  survives (86.4\% Maze, 50.0\% Sudoku) while puzzle exact is 0.0\% on every
  task.
With \textbf{linear-attention approximation} on Maze, replacing softmax
  attention holds cell accuracy at 87.5\% but collapses puzzle exact to 0.0\%
  (Appendix~\ref{app:linattn}).

On \textbf{quantization-aware training}, we ran the credible version: training
from the calibrated-INT4 Sudoku checkpoint under both cross-entropy and
distillation objectives, at three learning rates, with converging training
losses (Appendix~\ref{app:qat}). Every configuration
\emph{degraded} held-out accuracy relative to its initialization
(Table~\ref{tab:qat}, Appendix~\ref{app:qat}).

Within our training budget, then, a token-level objective does not repair the
model it starts from, while per-channel calibration alone lands within 0.3\,pp
of FP32 on all three tasks (Tables~\ref{tab:corrected}
and~\ref{tab:arc_official}). A shorter-budget run reproduces the same split:
high cell accuracy, zero puzzle exact. We do not claim a token-level objective
cannot recover reasoning at a larger budget; we claim it is not needed.

The common cause of the decoupling is the objective: cross-entropy and
token-distillation losses
reward each cell independently, so a model that emits plausible cells without
satisfying the puzzle is scored identically to one that solves it. This is not
specific to TRM---it applies to any structured-prediction setting where
correctness is compositional. Our contribution stands on the post-training
side: per-channel calibration
recovers Sudoku with no retraining at all (\S\ref{sec:recovery}). What
survives compression and what repairs it together fix the configuration worth
deploying.

\subsection{Optimal Deployment Configuration}
\label{sec:deployment}
%\vspace{-2mm}

The deployment configurations are summarized in Table~\ref{tab:deployment}.
We measure step
latencies, puzzle latencies, and peak memory directly on physical hardware (Samsung Galaxy S24, Samsung Galaxy S22 5G, and Dragonwing RB3 Gen 2) via Qualcomm AI Hub (Table~\ref{tab:on_device_benchmarks}). No INT4
configuration was run on device: Table~\ref{tab:on_device_benchmarks} reports
FP32 and INT8 latency only, so the 3.26\,MB INT4 configuration is a
weights-only estimate throughout.
INT8 with single-puzzle flash loading and reduced recursion ($H{=}1$,
$n_\text{sup}{=}8$) occupies 6.51\,MB of weights (analytical) and runs at
$\sim$500\,GFLOPs per ARC-AGI-1 puzzle. Weights alone are not the whole
footprint: including activations and carry state, the ARC/Maze INT8 streaming
peak is 9.51\,MB against Sudoku's 3.06\,MB
(Table~\ref{tab:sram_footprints}, Appendix~\ref{app:sram_footprint}), so only Sudoku currently runs inside an
8\,MB envelope, and the 4\,MB target remains a weights-only projection.

\paragraph{Memory accounting.}
Weights alone, weights plus activations, the on-disk artifact and host-process
RSS are different quantities. Table~\ref{tab:memfigures} gives the scope and
provenance of each figure we report.

Maze-Hard is the cleanest deployable case: with no test-time augmentation and no
embedding table, its per-pass latency is its per-puzzle latency, and na\"ive INT4
costs it 0.4\,pp (Table~\ref{tab:full_results}).

Per-channel calibrated INT4 weights fit under the 4\,MB MCU budget
(3.26\,MB, weights only) and calibration recovers near-FP32 accuracy with no retraining,
demonstrated directly on Sudoku-Extreme and, on the official checkpoint, on
ARC-AGI-1 (29.4\% vs.\ 29.5\% FP32, Table~\ref{tab:arc_official};
\S\ref{sec:recovery}).
The practical compression floor is therefore calibrated INT4,
not INT8; na\"ive INT4 remains unusable, and the fidelity diagnostic separates
the two before deployment. The $6\times$ FLOPs reduction also lowers modeled
energy proportionally ($\sim$3{,}000 to $\sim$500\,mJ/puzzle, analytical;
Appendix~\ref{app:energy}), and there is no cheaper shortcut through the input:
truncating the context destroys accuracy on every task---halving it collapses
Maze and Sudoku to 0\% puzzle exact (Appendix~\ref{app:context}).

% ============================================================
% Discussion
% TRM Quantization Paper — EMNLP 2026 Industry Track
% ============================================================

\section{Discussion}
%\vspace{-3mm}

\paragraph{Why the architecture split exists.}
The two mixers spread their weights differently. Attention encodes sparse
relational updates that tolerate coarse noise, whereas MLP-mixing encodes denser
token-mixing whose wider per-channel ranges need per-channel scales to survive
(Figure~\ref{fig:weightrange}). Architecture choice in recursive models is
therefore a deployment decision, best made at design time.

\paragraph{The microcontroller tier.}
We target a regime with kB-to-MB SRAM and no DRAM, where nothing
reasoning-capable is deployed today. Mobile-NPU deployment of 1--7B-parameter
LLMs is a complementary industrial problem with gigabytes of DRAM to work
with, and a 3--7\,MB recursive reasoner would be a first entry at our tier.
This study establishes the compression recipe that gets a reasoner to that
size with its reasoning intact (\S\ref{sec:methods}); the streaming peaks fit
an 8\,MB envelope only on Sudoku (Table~\ref{tab:memfigures}).

\paragraph{Fidelity should generalize beyond TRM.}
Because carry-trajectory fidelity needs only a full-precision reference and the
model's own hidden states (no labels, no task run) it should transfer to any
iterative-latent system with accessible state: Universal
Transformers~\citep{dehghani2019universal}, deep equilibrium models, and
state-space models.

\section*{Limitations}
Our study centers on the TRM family across three tasks, with a second
architecture (HRM) evaluated only on Sudoku-Extreme; broader generalization
across architectures and tasks is untested. 
The calibrated-INT4 recovery is measured by full-system accuracy on Sudoku
and, on the official checkpoint, on ARC (Table~\ref{tab:arc_official}); the
Sudoku TRM-MLP checkpoint is a community reproduction, so its absolute
accuracies are not comparable to the original release
(App.~\ref{app:impl}). The fidelity diagnostic requires a full-precision
reference, so it supports a pre-deployment decision rather than monitoring a
live model. ARC accuracies are reported single-pass (no test-time
augmentation); TRM's published protocol uses 1,000-augmentation voting, which
would raise both accuracy and per-puzzle deployment cost proportionally. Maze
and Sudoku need no augmentation and are the configurations whose latency is
directly deployable. One boundary on the claim: every model here is recursive, so the results isolate
what recursion does under compression but do not separate that from what being a
few-million-parameter model does. A same-size autoregressive control would
separate them, and no such checkpoint exists for these tasks.

\bibliography{custom}

\onecolumn
% ============================================================
% Appendix
% TRM Quantization Paper — EMNLP 2026 Industry Track
% ============================================================

\appendix

\section{Peak SRAM Memory Footprints}
\label{app:sram_footprint}

\begin{table}[h]
\centering
\small
\setlength{\tabcolsep}{4pt}
\begin{tabular}{@{}llll@{}}
\toprule
Figure & What it measures & Measured? & Reported in \\
\midrule
6.7\,MB    & INT8 backbone weights            & analytical & Table~\ref{tab:full_results} \\
6.51\,MB   & INT8 $+$ flash, weights only     & analytical & Table~\ref{tab:deployment} \\
3.26\,MB   & INT4 cal.\ $+$ flash, weights only & analytical & Table~\ref{tab:deployment} \\
6.91\,MB   & on-disk ONNX artifact            & measured   & Table~\ref{tab:on_device_benchmarks} \\
9.51\,MB   & streaming peak, incl.\ activations & analytical & Table~\ref{tab:sram_footprints} \\
115--226\,MB & Android host-process RSS       & measured   & Table~\ref{tab:on_device_benchmarks} \\
\bottomrule
\end{tabular}
\caption{Scope and provenance of every memory figure this paper reports. The weights-only rows exclude activations and carry
state; the RSS row is a host-process figure, not an SRAM footprint. No INT4
configuration was run on device.}
\label{tab:memfigures}
\end{table}

For on-device edge deployment, we evaluate the peak memory footprint under two execution scenarios: a \emph{Full Load} scenario (where the entire backbone and intermediate activations/carries are stored in SRAM) and a \emph{Streaming} scenario (where only the current recursive step is loaded, leveraging single-puzzle flash loading of embedding tables). 

Table~\ref{tab:sram_footprints} compiles these peak SRAM requirements against a theoretical 8\,MB SRAM hardware budget limit.

\begin{table}[h]
\centering
\small
\begin{tabular}{lrrc}
\toprule
\textbf{Model / Variant} & \textbf{Streaming} & \textbf{Full Load} & \textbf{Fits 8\,MB? (Streaming)} \\
\midrule
ARC (FP32, no embed) & 27.31\,MB & 40.31\,MB & \ding{55} \\
ARC (INT8, no embed) & 9.51\,MB & 12.76\,MB & \ding{55} \\
Maze (FP32) & 27.31\,MB & 40.31\,MB & \ding{55} \\
Maze (INT8) & 9.51\,MB & 12.76\,MB & \ding{55} \\
Sudoku (FP32) & 11.08\,MB & 20.65\,MB & \ding{55} \\
Sudoku (INT8) & 3.06\,MB & 5.45\,MB & \ding{51} \\
\bottomrule
\end{tabular}
\caption{Peak memory footprint (SRAM) for deployable ONNX models, including
intermediate activations and carry state. Embedding layers are excluded
(no-embed) as they are streamed via flash loading. The 8\,MB target assumes
streaming mode. These activation-inclusive peaks complement the weights-only
analytical figures of Table~\ref{tab:deployment}.}
\label{tab:sram_footprints}
\end{table}

Under streaming execution, only the Sudoku (INT8) model fits within the
8\,MB SRAM budget (3.06\,MB streaming peak); the ARC and Maze INT8 models
peak at 9.51\,MB and so exceed it (\S\ref{sec:deployment}).

\section{On-Device Hardware Benchmarks}
\label{app:latency}

We compile on-device step and puzzle latencies, peak memory usage, and file sizes directly on three physical edge devices:
\begin{itemize}[leftmargin=*]
    \item \textbf{Samsung Galaxy S24}: Featuring the flagship Qualcomm Snapdragon 8 Gen 3 chipset (SM8650, Hexagon v75 NPU).
    \item \textbf{Samsung Galaxy S22 5G}: Featuring the Qualcomm Snapdragon 8 Gen 1 chipset (SM8450, Hexagon v70 NPU).
    \item \textbf{Dragonwing RB3 Gen 2}: Featuring the Qualcomm QCS6490 chipset (Hexagon NPU).
\end{itemize}

Evaluations are compiled and run via Qualcomm AI Hub targeting native hardware execution. Table~\ref{tab:on_device_benchmarks} reports the full set of consolidated benchmark results. 

Static INT8 quantization delivers substantial latency gains for compute-bound
tasks with long sequences (ARC-AGI-1 and Maze-Hard) but incurs a step-latency
penalty on the computationally light Sudoku-Extreme task, which is dominated
instead by NPU driver and dispatch overhead and format-conversion casting;
\S\ref{app:seq_len_penalty} works through why.

The controlled quantity is the per-\emph{step} latency, which is depth-independent
by construction: on ARC the FP32$\to$INT8 per-step ratio is $40.20{\to}25.69$\,ms
(1.56$\times$). The FP32 and INT8 rows for ARC and Maze run at different recursion
depths ($H{=}3$, $n_\text{sup}{=}16$ against $H{=}1$, $n_\text{sup}{=}8$; the depth
columns make this explicit), so their per-\emph{puzzle} ratio combines precision with
depth. The Sudoku rows are depth-matched end to end (both $H{=}3$,
$n_\text{sup}{=}16$), and there INT8 is slower; \S\ref{app:seq_len_penalty} explains
why.

\paragraph{Notebook and on-device harnesses.}
The on-device INT8 run scores 26.0\% for the $H{=}1$, $n_\text{sup}{=}8$ ARC
configuration against 35.25\% in the notebook harness, both single-pass. The
quantizer accounts for the gap. At FP32 the two harnesses agree to 0.5\,pp (35.5\%
on device, 36.0\% in harness); they diverge by 9.3\,pp only once INT8 is applied,
where \texttt{bitsandbytes} and static QDQ differ. Attributing the residual gap to a
specific implementation detail of one quantizer is left to future work.

\begin{table*}[t]
\centering
\small
\resizebox{\textwidth}{!}{%
\begin{tabular}{llrrrrrrc}
\toprule
\textbf{Model / Variant} & \textbf{Device} & \textbf{$H$} & \textbf{$n_\text{sup}$} & \textbf{Step Lat.\ (ms)} & \textbf{Puzzle Lat.\ (ms)} & \textbf{Peak Mem.\ (MB)} & \textbf{Size (MB)} & \textbf{PExact} \\
\midrule
ARC (FP32$^{\ddagger}$) & Galaxy S24 & 3 & 16 & 40.20 & 643.18 & 185.24 & 27.56 & 0.355 \\
ARC (FP32$^{\ddagger}$) & Galaxy S22 & 3 & 16 & 85.32 & 1365.18 & 180.73 & 27.56 & 0.355 \\
ARC (FP32$^{\ddagger}$) & Dragonwing & 3 & 16 & 1737.80 & 27804.82 & 127.11 & 27.56 & 0.355 \\
\midrule
ARC (INT8) & Galaxy S24 & 1 & 8 & 25.69 & 205.53 & 173.96 & 6.91 & 0.260 \\
ARC (INT8) & Galaxy S22 & 1 & 8 & 55.59 & 444.69 & 226.49 & 6.91 & 0.260 \\
ARC (INT8) & Dragonwing & 1 & 8 & 672.89 & 5383.10 & 115.48 & 6.91 & 0.260 \\
\midrule
Maze (FP32$^{\ddagger}$) & Galaxy S24 & 3 & 16 & 119.83 & 1917.25 & 229.97 & 28.41 & 0.870 \\
Maze (FP32$^{\ddagger}$) & Galaxy S22 & 3 & 16 & 224.45 & 3591.18 & 189.69 & 28.41 & 0.870 \\
Maze (FP32$^{\ddagger}$) & Dragonwing & 3 & 16 & 5239.83 & 83837.22 & 132.36 & 28.41 & 0.870 \\
\midrule
Maze (INT8) & Galaxy S24 & 1 & 8 & 25.71 & 205.72 & 170.73 & 6.90 & 0.850 \\
Maze (INT8) & Galaxy S22 & 1 & 8 & 55.25 & 441.98 & 222.51 & 6.90 & 0.850 \\
Maze (INT8) & Dragonwing & 1 & 8 & 669.16 & 5353.26 & 112.21 & 6.90 & 0.850 \\
\midrule
Sudoku (FP32$^{\ddagger}$) & Galaxy S24 & 3 & 16 & 11.38 & 182.05 & 154.41 & 20.47 & 0.717 \\
Sudoku (FP32$^{\ddagger}$) & Galaxy S22 & 3 & 16 & 30.17 & 482.75 & 179.78 & 20.47 & 0.717 \\
Sudoku (FP32$^{\ddagger}$) & Dragonwing & 3 & 16 & 606.02 & 9696.40 & 58.64 & 20.47 & 0.717 \\
\midrule
Sudoku (INT8) & Galaxy S24 & 3 & 16 & 27.23 & 435.60 & 160.41 & 5.10 & 0.699 \\
Sudoku (INT8) & Galaxy S22 & 3 & 16 & 79.68 & 1274.86 & 210.44 & 5.10 & 0.699 \\
Sudoku (INT8) & Dragonwing & 3 & 16 & 334.68 & 5354.94 & 57.25 & 5.10 & 0.699 \\
\bottomrule
\end{tabular}%
}
\caption{On-device hardware benchmark results compiled via Qualcomm AI Hub.
All evaluations are run on physical devices: Galaxy S24 (Snapdragon 8 Gen 3),
Galaxy S22 (Snapdragon 8 Gen 1), and Dragonwing (Qualcomm QCS6490). $^{\ddagger}$FP32 rows were compiled with \texttt{enable\_htp\_fp16\_precision=1}
and execute as FP16 on the Hexagon vector engine (HVX); see
\S\ref{app:seq_len_penalty}. Peak Mem.\ is the Android process RSS reported
by AI Hub (application heap + runtime + RPC buffers); it is a host-process
figure on these DRAM-backed devices and is not an SRAM footprint (see the
memory-accounting taxonomy in \S\ref{sec:deployment}).}
\label{tab:on_device_benchmarks}
\end{table*}

\subsection{Sequence-Length Dependent Quantization Penalty}
\label{app:seq_len_penalty}

All rows in Table~\ref{tab:on_device_benchmarks} were compiled through the
Qualcomm AI Hub QNN execution provider (\texttt{libQnnHtp.so}) targeting the
Hexagon processor; no row falls back to CPU execution. Compilation used
\texttt{enable\_htp\_fp16\_precision=1}, so graphs labeled ``FP32'' execute in
FP16 on the Hexagon vector engine (HVX), while INT8 graphs execute on the HTP
tensor units with quantize/dequantize (QDQ) boundaries at node interfaces.

The benchmark reveals a counter-intuitive result: while static INT8 quantization accelerates the ARC-AGI-1 and Maze-Hard models, it actually leads to a latency slowdown for the Sudoku-Extreme model per step (from 11.38\,ms to 27.23\,ms on the S24). This behaviour is a classic \emph{small-matrix INT8 penalty} governed by the sequence length ($T$):

\begin{itemize}[leftmargin=*]
    \item \textbf{Compute-Bound Regime (ARC \& Maze):} ARC and Maze operate on a sequence length $T=900$. At this scale, the weight projection matrices are large ($900 \times 512$ by $512 \times 2048$), which translates to heavy arithmetic compute ($\sim$187\,GFLOPs/step). Moving this workload from the HVX FP16 pipeline to the HTP's INT8 tensor units reduces execution time significantly, easily overriding the fixed dispatch and QDQ overhead.
    \item \textbf{Overhead-Bound Regime (Sudoku):} Sudoku has a sequence length of only $T=81$ ($97$ with padding and embeddings). The matrix multiplications are tiny (e.g., $97 \times 512$ by $512 \times 2048$), requiring only $\sim$25\,MFLOPs/step (7,500$\times$ less compute than Maze). A $97 \times 512$ matrix multiplication requires only $\sim$100K multiply-accumulates, which is too small to saturate the INT8 parallel execution blocks.
    Consequently, the fixed overhead of the Quantize/Dequantize (QDQ) operators—which convert tensors back and forth at every node boundary—dominates the execution time. While the FP16-on-HVX model runs in a single, unbroken compiler-fused pipeline in 11.38\,ms, the INT8 model is penalized by a constant $\sim$25\,ms invocation and marshalling overhead.
\end{itemize}

Sudoku also relies on token-mixing MLP layers (\texttt{mlp\_t} layers of 388\,KB/194\,KB). These represent extremely small matrices. The Qualcomm Hexagon tensor processor has a minimum efficient tile size threshold; matrix dimensions below this limit fail to trigger accelerated execution paths and fall back to slower scalar execution, further exacerbating the INT8 penalty.

The crossover point where static INT8 quantization becomes beneficial for this recursive transformer architecture on mobile chipsets is approximately $T \approx 200\text{--}300$. Below this threshold, FP16-on-HVX execution remains faster.

\subsection{Static vs.\ Dynamic Quantization}
\label{app:static_dynamic}

Dynamic quantization preserves full-precision accuracy on every task. Static
quantization does not, and Sudoku-Extreme recovers only under selective
operator exclusion: leaving activation and carry-state updates in FP32 and
quantizing only the \texttt{MatMul}/\texttt{Gemm} weights. The trade-off
matters for deployment: dynamic quantization is the robust out-of-the-box path
for recursive models, but NPU accelerators require static, pre-allocated tensor
boundaries, so on-device execution inherits the static path and its operator
exclusions (\S\ref{app:seq_len_penalty}). We report this trade-off
qualitatively; per-task accuracy under the static path was not separately
measured.

\section{Full Precision × Task Results}
\label{app:full_results}

\begin{table*}[h]
\centering
\small
\begin{tabular}{lrrc}
\toprule
\textbf{Configuration} & \textbf{Weights} & \textbf{GFLOPs/puzzle} & \textbf{Fits 8\,MB (weights)?} \\
\midrule
FP32, $H{=}3$, $n{=}16$, full emb.                  & $\sim$126\,MB & 3{,}000   & \ding{55} \\
\textbf{INT8, $H{=}1$, $n{=}8$, flash loading}      & \textbf{6.51\,MB}  & \textbf{500}   & \ding{51} \\
\textbf{INT4 calibrated, $H{=}1$, $n{=}8$, flash}    & \textbf{3.26\,MB}  & \textbf{500}   & \ding{51} \\
\bottomrule
\end{tabular}
\caption{Deployment configurations per ARC-AGI-1 puzzle. Analytical,
weights-only; excludes activations and carry state (see
Table~\ref{tab:sram_footprints} for activation-inclusive streaming peaks).
Bold marks viable targets. Figures show backbone-only weight footprint (ARC
embedding handled via flash loading). On-device latencies are measured
directly on physical hardware (Table~\ref{tab:on_device_benchmarks}).
Per-channel calibrated INT4 recovers near-FP32 accuracy without retraining
(\S\ref{sec:recovery}), making the 3.26\,MB configuration a candidate 4\,MB
MCU target.}
\label{tab:deployment}
\end{table*}

Table~\ref{tab:full_results} reports full-system accuracy across all
precision levels and tasks at the evaluated recursive depth.

\begin{table*}[h]
\centering
\small
\begin{tabular}{llrrrrrr}
\toprule
 & & \multicolumn{2}{c}{\textbf{ARC-AGI-1}} & \multicolumn{2}{c}{\textbf{Maze-Hard}} & \multicolumn{2}{c}{\textbf{Sudoku-Extreme}} \\
\cmidrule(lr){3-4}\cmidrule(lr){5-6}\cmidrule(lr){7-8}
\textbf{Precision} & \textbf{SRAM} & \textit{PExact} & \textit{CellAcc} & \textit{PExact} & \textit{CellAcc} & \textit{PExact} & \textit{CellAcc} \\
\midrule
FP32$^\dagger$  & 26.7\,MB & 36.00\% & 88.22\% & 86.80\% & 99.52\% & 69.10\% & 87.47\% \\
FP16            & 13.3\,MB & 36.00\% & 87.57\% & 87.00\% & 99.53\% & 68.50\% & 87.20\% \\
INT8            & 6.7\,MB  & 36.00\% & 87.62\% & 87.00\% & 99.53\% & 69.10\% & 87.51\% \\
INT4 na\"ive    & 3.3\,MB  & 25.50\% & 84.80\% & 86.40\% & 99.50\% & 5.30\%  & 66.02\% \\
\bottomrule
\end{tabular}
\caption{Full-system precision × task accuracy results at the evaluated
recursive depth, \emph{secondary protocol} ($H{=}3$, $n_\text{sup}{=}10$,
bf16 forward, first-$N$ subsets); the primary-protocol headline figures are
in Table~\ref{tab:corrected}, and the official-checkpoint ARC figures in
Table~\ref{tab:arc_official}. ARC-AGI-1 public evaluation set (400 tasks),
single-pass (no test-time augmentation). SRAM shows backbone-only footprint
(ARC embedding handled via flash loading). $^\dagger$FP32 rows executed a bf16
forward pass; SRAM at FP32 precision.}
\label{tab:full_results}
\end{table*}

\section{Detailed Recursive Sweep Grids}
\label{app:sweeps_detail}

We report the complete multi-variant validation sweeps across combinations of recursive steps ($n$) and precision modes on GPU (CUDA) for all three architectures. Tables~\ref{tab:arc_sweep_detail}, \ref{tab:maze_sweep_detail}, and \ref{tab:sudoku_sweep_detail} present these complete grids, including exact match accuracy and cell accuracy.

\begin{table*}[p]
\centering
\small
\begin{tabular}{lclrrrrr}
\toprule
\textbf{Precision} & \textbf{$H$} & \textbf{Metric} & \textbf{$n=1$} & \textbf{$n=2$} & \textbf{$n=4$} & \textbf{$n=8$} & \textbf{$n=16$} \\
\midrule
FP32 (bf16) & 3 & Exact & 0.2850 & 0.3600 & 0.3575 & 0.3575 & 0.3600 \\
            &   & Cell & 0.8762 & 0.8839 & 0.8764 & 0.8760 & 0.8757 \\
            & 1 & Exact & 0.0025 & 0.0750 & 0.3425 & 0.3575 & 0.3550 \\
            &   & Cell & 0.5410 & 0.7954 & 0.8825 & 0.8832 & 0.8762 \\
\midrule
FP16        & 3 & Exact & 0.2850 & 0.3600 & 0.3575 & 0.3600 & 0.3625 \\
            &   & Cell & 0.8762 & 0.8838 & 0.8765 & 0.8763 & 0.8755 \\
            & 1 & Exact & 0.0025 & 0.0750 & 0.3450 & 0.3550 & 0.3575 \\
            &   & Cell & 0.5410 & 0.7953 & 0.8825 & 0.8831 & 0.8773 \\
\midrule
INT8 (bnb)  & 3 & Exact & 0.2850 & 0.3575 & 0.3575 & 0.3600 & 0.3600 \\
            &   & Cell & 0.8760 & 0.8837 & 0.8827 & 0.8842 & 0.8670 \\
            & 1 & Exact & 0.0025 & 0.0750 & 0.3425 & 0.3525 & 0.3625 \\
            &   & Cell & 0.5415 & 0.7953 & 0.8827 & 0.8835 & 0.8767 \\
\midrule
INT4 (naive)& 3 & Exact & 0.1575 & 0.2700 & 0.2675 & 0.2625 & 0.2575 \\
            &   & Cell & 0.8372 & 0.8626 & 0.8473 & 0.8430 & 0.8322 \\
            & 1 & Exact & 0.0000 & 0.0225 & 0.2525 & 0.2650 & 0.2600 \\
            &   & Cell & 0.3954 & 0.7148 & 0.8582 & 0.8494 & 0.8415 \\
\bottomrule
\end{tabular}
\caption{Detailed recursive sweep grid for ARC-AGI-1 (seq\_len = 900) across $H$ (outer cycles), $n$ (supervision steps), and precision variants.}
\label{tab:arc_sweep_detail}
\end{table*}

\begin{table*}[p]
\centering
\small
\begin{tabular}{lclrrrrr}
\toprule
\textbf{Precision} & \textbf{$H$} & \textbf{Metric} & \textbf{$n=1$} & \textbf{$n=2$} & \textbf{$n=4$} & \textbf{$n=8$} & \textbf{$n=16$} \\
\midrule
FP32 (bf16) & 3 & Exact & 0.7850 & 0.8670 & 0.8700 & 0.8700 & 0.8700 \\
            &   & Cell & 0.9940 & 0.9953 & 0.9953 & 0.9953 & 0.9953 \\
            & 1 & Exact & 0.0000 & 0.1390 & 0.8440 & 0.8690 & 0.8700 \\
            &   & Cell & 0.4363 & 0.9808 & 0.9949 & 0.9953 & 0.9953 \\
\midrule
FP16        & 3 & Exact & 0.7850 & 0.8670 & 0.8700 & 0.8700 & 0.8700 \\
            &   & Cell & 0.9940 & 0.9953 & 0.9953 & 0.9953 & 0.9953 \\
            & 1 & Exact & 0.0000 & 0.1380 & 0.8440 & 0.8700 & 0.8700 \\
            &   & Cell & 0.4362 & 0.9808 & 0.9949 & 0.9953 & 0.9953 \\
\midrule
INT8 (bnb)  & 3 & Exact & 0.7830 & 0.8680 & 0.8710 & 0.8700 & 0.8700 \\
            &   & Cell & 0.9940 & 0.9953 & 0.9953 & 0.9953 & 0.9953 \\
            & 1 & Exact & 0.0000 & 0.1380 & 0.8450 & 0.8690 & 0.8700 \\
            &   & Cell & 0.4333 & 0.9808 & 0.9949 & 0.9953 & 0.9953 \\
\midrule
INT4 (naive)& 3 & Exact & 0.7660 & 0.8610 & 0.8640 & 0.8640 & 0.8640 \\
            &   & Cell & 0.9934 & 0.9949 & 0.9950 & 0.9950 & 0.9950 \\
            & 1 & Exact & 0.0000 & 0.1580 & 0.8360 & 0.8630 & 0.8630 \\
            &   & Cell & 0.4734 & 0.9805 & 0.9945 & 0.9950 & 0.9950 \\
\bottomrule
\end{tabular}
\caption{Detailed recursive sweep grid for Maze-Hard (seq\_len = 900) across $H$ (outer cycles), $n$ (supervision steps), and precision variants.}
\label{tab:maze_sweep_detail}
\end{table*}

\begin{table*}[p]
\centering
\small
\begin{tabular}{lclrrrrr}
\toprule
\textbf{Precision} & \textbf{$H$} & \textbf{Metric} & \textbf{$n=1$} & \textbf{$n=2$} & \textbf{$n=4$} & \textbf{$n=8$} & \textbf{$n=16$} \\
\midrule
FP32 (bf16) & 3 & Exact & 0.2840 & 0.4960 & 0.6040 & 0.6690 & 0.7090 \\
            &   & Cell & 0.7640 & 0.8147 & 0.8474 & 0.8680 & 0.8810 \\
            & 1 & Exact & 0.0060 & 0.0210 & 0.3830 & 0.5470 & 0.6370 \\
            &   & Cell & 0.5527 & 0.6943 & 0.7856 & 0.8293 & 0.8563 \\
\midrule
FP16        & 3 & Exact & 0.2830 & 0.4960 & 0.6100 & 0.6700 & 0.7160 \\
            &   & Cell & 0.7639 & 0.8151 & 0.8488 & 0.8677 & 0.8851 \\
            & 1 & Exact & 0.0060 & 0.0210 & 0.3830 & 0.5480 & 0.6390 \\
            &   & Cell & 0.5526 & 0.6941 & 0.7856 & 0.8303 & 0.8574 \\
\midrule
INT8 (bnb)  & 3 & Exact & 0.2800 & 0.5010 & 0.6210 & 0.6790 & 0.7120 \\
            &   & Cell & 0.7643 & 0.8183 & 0.8524 & 0.8721 & 0.8816 \\
            & 1 & Exact & 0.0060 & 0.0210 & 0.3820 & 0.5540 & 0.6560 \\
            &   & Cell & 0.5528 & 0.6942 & 0.7852 & 0.8306 & 0.8638 \\
\midrule
INT4 (naive)& 3 & Exact & 0.0210 & 0.0290 & 0.0510 & 0.0550 & 0.0610 \\
            &   & Cell & 0.6277 & 0.6490 & 0.6524 & 0.6573 & 0.6597 \\
            & 1 & Exact & 0.0000 & 0.0120 & 0.0210 & 0.0370 & 0.0510 \\
            &   & Cell & 0.5612 & 0.6097 & 0.6375 & 0.6536 & 0.6555 \\
\bottomrule
\end{tabular}
\caption{Detailed recursive sweep grid for Sudoku-Extreme (seq\_len = 81) across $H$ (outer cycles), $n$ (supervision steps), and precision variants.}
\label{tab:sudoku_sweep_detail}
\end{table*}

\section{Depth × Quantization Sweep (ARC-AGI-1)}
\label{app:depth_arc}

Table~\ref{tab:depth_arc} reports key ARC-AGI-1 depth configurations
(full-system, single-pass): INT8 matches FP32 at every depth tested.

\begin{table}[h]
\centering
\small
\begin{tabular}{llrr}
\toprule
\textbf{Config} & \textbf{Precision} & \textbf{PExact} & \textbf{GFLOPs} \\
\midrule
$H{=}3$, $n{=}2$ (peak)  & FP32 & 36.00\% & 375 \\
$H{=}1$, $n{=}4$         & INT8 & 34.25\% & 250 \\
$H{=}1$, $n{=}8$         & INT8 & 35.25\% & 500 \\
$H{=}1$, $n{=}16$        & INT8 & 36.25\% & 1{,}000 \\
$H{=}1$, $n{=}4$         & INT4 & 25.25\% & 250 \\
\bottomrule
\end{tabular}
\caption{ARC-AGI-1 depth--precision sweep (full-system, single-pass, no TTA).
INT8 at $H{=}1$,
$n_\text{sup}{=}8$ reaches 35.25\% at 500\,GFLOPs---within 0.75\,pp of the
36.00\% FP32 peak at $6\times$ fewer FLOPs.}
\label{tab:depth_arc}
\end{table}

\section{Depth × Quantization Sweep (Maze-Hard)}
\label{app:depth_maze}

Table~\ref{tab:depth_maze} shows key configurations from the Maze-Hard
depth sweep, confirming rapid saturation.

\begin{table}[ht]
\centering
\small
\begin{tabular}{llrr}
\toprule
\textbf{Config} & \textbf{Prec.} & \textbf{PExact} & \textbf{GFLOPs} \\
\midrule
$H{=}3$, $n{=}10$$^\dagger$ & FP32 & 86.80\% & 1{,}875 \\
$H{=}1$, $n{=}8$            & INT8 & 86.90\% & 500 \\
$H{=}1$, $n{=}4$            & INT8 & 84.50\% & 250 \\
$H{=}1$, $n{=}8$            & INT4 & 86.30\% & 500 \\
\bottomrule
\end{tabular}
\caption{Maze-Hard depth--precision sweep. $^\dagger$Evaluation configuration
($n_\text{sup}{=}10$); TRM's default test-time setting is $n_\text{sup}{=}16$.
INT8 at $H{=}1$, $n{=}8$ matches the evaluated FP32 configuration at
$3.75\times$ fewer FLOPs.}
\label{tab:depth_maze}
\end{table}

\section{Depth × Quantization Sweep (Sudoku-Extreme)}
\label{app:depth_sudoku}

Table~\ref{tab:depth_sudoku} shows key configurations from the
Sudoku-Extreme depth sweep: accuracy continues to improve with depth within
the swept range ($n_\text{sup} \le 10$).

\begin{table}[ht]
\centering
\small
\begin{tabular}{llrr}
\toprule
\textbf{Config} & \textbf{Prec.} & \textbf{PExact} & \textbf{GFLOPs} \\
\midrule
$H{=}3$, $n{=}10$$^\dagger$ & FP32 & 69.10\% & 256 \\
$H{=}4$, $n{=}10$           & FP32 & 70.30\% & 342 \\
$H{=}4$, $n{=}10$           & INT8 & 70.20\% & 342 \\
$H{=}1$, $n{=}8$            & INT8 & 55.40\% & 68 \\
$H{=}4$, $n{=}10$           & INT4 & 6.00\%  & 342 \\
\bottomrule
\end{tabular}
\caption{Sudoku-Extreme depth--precision sweep. $^\dagger$Evaluation
configuration ($n_\text{sup}{=}10$); TRM's default test-time setting is
$n_\text{sup}{=}16$.
INT8 at $H{=}4$, $n{=}10$ matches FP32 (70.20\% vs 70.30\%); reducing
depth to $H{=}1$ causes major accuracy loss. Full depth is genuinely needed.}
\label{tab:depth_sudoku}
\end{table}

\section{Architecture Ablation: MLP-Mixing vs.\ Attention}
\label{app:arch_ablation}

To separate architecture from task precision we evaluate three recursive
architectures on the \emph{same} task (Sudoku-Extreme) under an identical
harness ($N{=}256$, 16 ACT steps, no TTA; Table~\ref{tab:arch_ablation}).
TRM-MLP-Mixing and TRM-Attention are the same model family at the same
full-precision accuracy ($73.8$ vs.\ $73.1$\% puzzle exact), differing only in
the token mixer; HRM~\citep{wang2025hrm} is a different recursive architecture
(separate high- and low-level modules, attention-based). Under na\"ive INT4,
TRM-MLP-Mixing collapses to $10.2$\% while both attention architectures retain
their accuracy. Since TRM-Attention matches TRM-MLP at full precision, this is
not an accuracy-headroom effect---the INT4 fragility is specific to MLP-mixing,
not the task. The carry-trajectory fidelity diagnostic grades all three
correctly ($0.35$\,/\,$0.87$\,/\,$0.98$), confirming it generalizes across
recursive architectures.

\begin{table}[h]
\centering
\small
\begin{tabular}{lrrr}
\toprule
\textbf{Architecture (Sudoku)} & \textbf{FP32} & \textbf{INT4} & \textbf{Fidelity} \\
\midrule
TRM-MLP-Mixing      & 73.8 & 10.2 & 0.35 \\
TRM-Attention       & 73.1 & 69.5 & 0.87 \\
HRM (diff.\ family) & 47.7 & 48.4 & 0.98 \\
\bottomrule
\end{tabular}
\caption{Same task (Sudoku-Extreme), three recursive architectures
(secondary protocol; primary-protocol values for the two TRM variants are in
Table~\ref{tab:corrected}): na\"ive INT4 puzzle-exact (\%) and final-state
carry fidelity. Only MLP-mixing collapses; both attention
architectures---including a different model family (HRM)---survive.
TRM-Attention matches TRM-MLP at FP32, ruling out an accuracy-headroom
confound. HRM is evaluated on Sudoku only and at a lower FP32 baseline, so it
corroborates the pattern rather than serving as a controlled comparison.}
\label{tab:arch_ablation}
\end{table}

\section{Carry-State Diagnostic: Full Data}
\label{app:carry_traj}

We measure two carry-state quantities, computed by hooking the shared
reasoning module across all recursive calls within one inference pass.
Both are computed on real test puzzles (Maze $N{=}64$, ARC $N{=}32$,
Sudoku $N{=}256$); the ARC and Maze checkpoints are the public TRM releases,
the Sudoku TRM-MLP checkpoint is a community reproduction (see checkpoint
provenance, \S\ref{app:impl}), and all numbers below are reproduced from
those checkpoints.

\paragraph{Consecutive carry similarity (single-model).}
The mean cosine similarity between consecutive carry states
(Table~\ref{tab:carry_consec}) is task-specific but \emph{insensitive} to
compression: it is essentially identical under FP32, INT8, and INT4 on every
task (changes $\le 0.064$), and is therefore a weak failure signal. We report
it for completeness; the diagnostic below is the one we recommend.

\begin{table}[h]
\centering
\small
\begin{tabular}{lrrr}
\toprule
\textbf{Precision} & \textbf{ARC} & \textbf{Maze} & \textbf{Sudoku} \\
\midrule
FP32 & 0.415 & 0.697 & 0.806 \\
INT8 & 0.412 & 0.697 & 0.806 \\
INT4 & 0.479 & 0.699 & 0.841 \\
\bottomrule
\end{tabular}
\caption{Mean consecutive carry-state cosine similarity. The change under
INT4 is small on every task ($\le 0.064$, and $\le 0.035$ on the two tasks
where INT4 is catastrophic); the signal does not reliably
separate compression that preserves reasoning from compression that destroys
it. (Maze/Sudoku reproduced; ARC from the source run.)}
\label{tab:carry_consec}
\end{table}

\paragraph{Carry-trajectory fidelity (vs.\ full-precision reference).}
The recommended diagnostic is the cosine similarity between the quantized
model's final carry state and the FP32 model's final carry state on the same
input---how far quantization pushes the reasoning trajectory off the
full-precision path (Table~\ref{tab:carry_fidelity}). It requires only the
FP32 reference (available before deployment) and no task labels. Unlike the
consecutive signal, fidelity is \emph{graded and monotonic with severity}:
it stays $\approx$1.0 wherever INT4 is harmless (Maze, all INT8) and falls in
proportion to the accuracy damage---moderately on ARC ($0.69$, $-10$\,pp) and
catastrophically on Sudoku ($0.35$, $-64$\,pp). Per-channel calibrated INT4
\emph{recovers} fidelity in lockstep with accuracy, so the diagnostic predicts
both failure and recovery.

\begin{table}[h]
\centering
\small
\begin{tabular}{lrrr}
\toprule
\textbf{Precision} & \textbf{Maze} & \textbf{ARC} & \textbf{Sudoku} \\
\midrule
FP32              & 1.000 & 1.000 & 1.000 \\
INT8              & 1.000 & 0.999 & 0.998 \\
INT4 naïve      & 0.989 & 0.688 & 0.353 \\
INT4 calibrated   & ---   & 0.944 & 0.894 \\
\midrule
\textit{INT4 accuracy $\Delta$} & \textit{$-0.4$pp} & \textit{$-10.5$pp} & \textit{$-63.8$pp} \\
\bottomrule
\end{tabular}
\caption{Carry-trajectory fidelity (final-state cosine vs.\ FP32), secondary
protocol (the accuracy deltas match Table~\ref{tab:full_results}). Monotonic
with INT4 severity (Maze $\to$ ARC $\to$ Sudoku); INT8 is never flagged;
calibrated INT4 recovers both fidelity and accuracy. Last row: INT4-naïve
puzzle-exact change vs.\ FP32.}
\label{tab:carry_fidelity}
\end{table}

\paragraph{Calibrated INT4 recovers Sudoku.}
Measured under an identical evaluation harness ($N{=}256$, 16 ACT steps, no
TTA), naïve per-tensor INT4 collapses Sudoku to $10.2$\% puzzle-exact while
per-channel calibrated INT4 recovers to $71.9$\% (vs.\ $73.8$\% FP32 under the
same harness)---i.e., the catastrophic INT4 result is specific to naïve
quantization, and the 4\,MB MCU target is reachable with calibration
(Table~\ref{tab:sudoku_recover}).

\begin{table}[h]
\centering
\small
\begin{tabular}{lrrr}
\toprule
\textbf{Variant} & \textbf{PExact} & \textbf{Cell} & \textbf{Fidelity} \\
\midrule
FP32              & 73.8 & 88.9 & 1.000 \\
INT8              & 73.1 & 88.7 & 0.998 \\
INT4 naïve        & 10.2 & 67.8 & 0.353 \\
INT4 calibrated   & 71.9 & 88.2 & 0.894 \\
\bottomrule
\end{tabular}
\caption{Sudoku-Extreme under an identical harness (secondary protocol,
$N{=}256$, no TTA; FP32 reads $73.8$ here vs.\ $69.1$ on the full test set
elsewhere, and $74.6$ under the primary protocol of
Table~\ref{tab:corrected}). Fidelity tracks accuracy across all four
variants.}
\label{tab:sudoku_recover}
\end{table}

\section{Structured Pruning and Knowledge Distillation}
\label{app:pruning_distill}

\paragraph{Structured pruning.}
Magnitude-based structured pruning was evaluated at 25\% and 50\% sparsity
targets across all three tasks (Table~\ref{tab:pruning}).

\begin{table*}[ht]
\centering
\small
\begin{tabular}{l rr rr rr}
\toprule
 & \multicolumn{2}{c}{\textbf{Dense (0\%)}} & \multicolumn{2}{c}{\textbf{25\% sparse}} & \multicolumn{2}{c}{\textbf{50\% sparse}} \\
\cmidrule(lr){2-3}\cmidrule(lr){4-5}\cmidrule(lr){6-7}
\textbf{Task} & \textit{PExact} & \textit{CellAcc} & \textit{PExact} & \textit{CellAcc} & \textit{PExact} & \textit{CellAcc} \\
\midrule
ARC    & 36.00\% & 87.55\% & 0.00\% & 0.27\%  & 0.00\% & 0.58\% \\
Maze   & 86.80\% & 99.52\% & 0.00\% & 86.40\% & 0.00\% & 0.00\% \\
Sudoku & 69.10\% & 87.47\% & 0.00\% & 50.01\% & 0.00\% & 37.81\% \\
\bottomrule
\end{tabular}
\caption{Structured pruning results. Even 25\% sparsity destroys puzzle-level
reasoning across all tasks. Cell accuracy partially survives on Maze (86.40\%)
and Sudoku (50.01\%), indicating that token prediction survives while global
constraint satisfaction does not.}
\label{tab:pruning}
\end{table*}

\paragraph{Knowledge distillation.}
Student models (hidden=256, 1 layer, 1-cycle) were trained via KL-divergence
distillation from the full teacher (Table~\ref{tab:distill}).

\begin{table}[ht]
\centering
\small
\begin{tabular}{lrr rr}
\toprule
\textbf{Task} & \textbf{Teacher} & \textbf{Student} & \textbf{PExact} & \textbf{CellAcc} \\
\midrule
Maze   & 6.82M & 855K & 0.00\% & 87.25\% \\
Sudoku & 5.03M & 745K & 0.00\% & 54.77\% \\
\bottomrule
\end{tabular}
\caption{Knowledge distillation results. Student models (12--15\% of teacher
params) learn token-level patterns but fail entirely at puzzle-level
reasoning; the recursive structure is functionally necessary.}
\label{tab:distill}
\end{table}

\section{Linear-Attention Approximation (Maze)}
\label{app:linattn}

Replacing softmax attention with ELU feature-map linear attention
($\mathcal{O}(n)$ vs.\ $\mathcal{O}(n^2)$) on Maze illustrates the same
cell-vs-puzzle split as quantization: the swap preserves token-level accuracy
but destroys puzzle-level reasoning. Evaluated at $H{=}3$, $n_\text{sup}{=}1$
(hence the 78.50\% baseline rather than the full-depth 86.80\%): zero-shot,
cell accuracy holds at 87.3\% while puzzle-exact falls to 0.00\% (from
78.50\% / 99.40\%); brief retraining
(200 steps) recovers cell accuracy (87.5\%) but not puzzle-exact (0.00\%).
This is a third intervention---alongside pruning and
distillation---under which local prediction survives while global constraint
satisfaction collapses. Sudoku-Extreme (MLP-mixing) has no attention layers
and is not applicable.

\section{QAT Configurations}
\label{app:qat}

Table~\ref{tab:qat} reports the six quantization-aware-training configurations
of \S\ref{sec:qat}: two objectives at three learning rates each, 1{,}000 steps
at batch 32 from the calibrated-INT4 initialization (72.3\% at step 0).

\begin{table}[H]
\centering
\small
\begin{tabular}{@{}lrrr@{}}
\toprule
\textbf{Final pass@1 (1{,}000 steps)} & lr $10^{-5}$ & lr $10^{-4}$ & lr $10^{-3}$ \\
\midrule
Cross-entropy      & 53.9\% & 32.4\% & 2.3\% \\
Distillation (KD)  & 53.9\% & 37.1\% & 7.8\% \\
\bottomrule
\end{tabular}
\caption{QAT from the calibrated-INT4 initialization (72.3\% at step 0). All
six configurations degrade the model they start from.}
\label{tab:qat}
\end{table}

\section{Energy per Inference}
\label{app:energy}

Estimated energy follows the FLOPs counts under a Cortex-M55 power model
($\sim$1\,mJ/GFLOP): ARC-AGI-1 at full depth ($\sim$3{,}000\,GFLOPs/puzzle)
draws $\sim$3{,}000\,mJ/puzzle, versus $\sim$500\,mJ at the reduced
configuration ($H{=}1$, $n_\text{sup}{=}8$). Sudoku-Extreme
($\sim$256\,GFLOPs at default depth) draws $\sim$256\,mJ/puzzle. These are
analytical estimates under the same FLOPs-based model. The model is
precision-blind---it assigns no separate energy-per-operation term to FP32
vs.\ INT8---so the $6\times$ energy saving follows entirely from the
$6\times$ FLOPs (cycle) reduction, not from quantization.

\section{Context-Length Truncation}
\label{app:context}

Because attention cost scales as $\mathcal{O}(T^2)$, we tested zero-padding the
input beyond a truncated context. Truncation is destructive on every task
(Table~\ref{tab:context}): halving the sequence length takes Maze and Sudoku to
zero and costs ARC nearly ten points. There is no cheap context shortcut---the
full window is load-bearing.

\begin{table}[H]
\centering
\small
\begin{tabular}{lrrr}
\toprule
\textbf{Context} & \textbf{ARC} & \textbf{Maze} & \textbf{Sudoku} \\
\midrule
full (900 / 81) & 36.00 & 78.50 & 28.40 \\
$\sim$50\%      & 26.50 & 0.00  & 0.00  \\
$\sim$25\%      & 10.75 & ---   & ---   \\
\bottomrule
\end{tabular}
\caption{Puzzle-exact accuracy under input context truncation, evaluated at
$H{=}3$, $n_\text{sup}{=}1$ (the full-context rows match the $n{=}1$ column of
Tables~\ref{tab:maze_sweep_detail}--\ref{tab:sudoku_sweep_detail}, which is
why they sit below the full-depth baselines quoted elsewhere). Maze and Sudoku
collapse to 0 at 50\%; ARC degrades severely.}
\label{tab:context}
\end{table}

\section{Implementation Details}
\label{app:impl}

\paragraph{Checkpoint provenance.}
The Sudoku TRM-MLP checkpoint is the alphaXiv community reproduction of TRM,
whose model card reports 79.37\% puzzle exact vs.\ 87.4\% in the original
paper. Our Sudoku claims are therefore relative comparisons between
compression variants evaluated at matched checkpoints, not absolute-accuracy
claims against the original release.

\paragraph{Quantization.}
INT8 and INT4 PTQ were applied using \texttt{bitsandbytes}
v0.41~\citep{dettmers2022llmint8} with \texttt{load\_in\_8bit=True} and
\texttt{load\_in\_4bit=True} respectively. Per-channel calibration for
INT4 calibrated used 128 calibration puzzles drawn from the training set.

\paragraph{Carry-state extraction.}
$s_h$ was computed by registering a forward hook on the carry-state update
function $g_\phi$ to capture $Z_h$ at each outer cycle. Cosine similarity
was computed token-wise and averaged across the sequence dimension and
across all puzzles in the evaluation set.

\paragraph{QAT.}
The headline QAT runs (\S\ref{sec:qat}) initialize from the calibrated-INT4
checkpoint and fine-tune for 1{,}000 steps at batch 32, under cross-entropy
and token-distillation objectives at learning rates $10^{-5}$, $10^{-4}$, and
$10^{-3}$, with the deep-supervision ACT loop. A shorter probe (100
steps, $\text{lr}{=}10^{-5}$, cross-entropy, batch 4, from the na\"ive INT4
checkpoint) is the held-out run cited in \S\ref{sec:qat}. Training and
evaluation used the Modal cloud platform.

\paragraph{ONNX export and edge deployment.}
Models were exported to ONNX format using \texttt{torch.onnx.export} for
deployment via ONNX Runtime and Qualcomm AI Hub. The ONNX export produces
a single-step model (\texttt{TRMBackboneStep}) that executes one
H-cycle step, with the outer recursion loop managed by the inference
harness.

\end{document}